\documentclass[10pt,journal,compsoc]{IEEEtran}

\ifCLASSOPTIONcompsoc
  \usepackage[nocompress]{cite}
  \usepackage{array,multirow,graphicx}
  \usepackage{tabularx,booktabs,ragged2e}
  \usepackage{amsmath}
\else
  \usepackage{cite}
\fi
\usepackage{comment}

\begin{document}

\title{DeepFire2: A Convolutional Spiking Neural Network Accelerator on FPGAs}

\author{Myat~Thu~Linn~Aung,~ Daniel~Gerlinghoff,~ Chuping~Qu,~ Liwei~Yang,~ Tian~Huang,\\ Rick~Siow~Mong~Goh,~ Tao~Luo,~ Weng-Fai~Wong%
\IEEEcompsocitemizethanks{
    \IEEEcompsocthanksitem M.T.L.~Aung, D.~Gerlinghoff, C.~Qu, L.~Yang, T.~Huang, R.S.M.~Goh, and T. Luo are with the Institute of High Performance Computing, Agency for Science Technology and Research (A*STAR), Singapore.
    \IEEEcompsocthanksitem W.F.~Wong is with the Department of Computer Science, National University of Singapore.
    \IEEEcompsocthanksitem Correspondence to T.~Luo (leto.luo@gmail.com)}
}

\IEEEtitleabstractindextext{%
\begin{abstract}
    Brain-inspired spiking neural networks (SNNs) replace the multiply-accumulate operations of traditional neural networks by integrate-and-fire neurons, with the goal of achieving greater energy efficiency. Specialized hardware implementations of those neurons clearly have advantages over general-purpose devices in terms of power and performance, but exhibit poor scalability when it comes to accelerating large neural networks. DeepFire2 introduces a hardware architecture which can map large network layers efficiently across multiple super logic regions in a multi-die FPGA. That gives more control over resource allocation and parallelism, benefiting both throughput and energy consumption. Avoiding the use of lookup tables to implement the \texttt{AND} operations of an SNN, prevents the layer size to be limited by logic resources. A deep pipeline does not only lead to an increased clock speed of up to 600~MHz. We double the throughput and power efficiency compared to our previous version of DeepFire, which equates to an almost 10-fold improvement over other previous implementations. Importantly, we are able to deploy a large ImageNet model, while maintaining a throughput of over 1500 frames per second.
\end{abstract}

}

\newcommand{\IEEEcopyright}{
\noindent\parbox{\textwidth}{%
    \vspace{\baselineskip}
    \textcopyright~2023 IEEE. Personal use of this material is permitted. Permission from IEEE must be obtained for all other uses, in any current or future media, including reprinting/republishing this material for advertising or promotional purposes, creating new collective works, for resale or redistribution to servers or lists, or reuse of any copyrighted component of this work in other works.
    
    \vspace{\baselineskip}
    DOI: 10.1109/TC.2023.3272284}
}

\IEEEcopyright

\maketitle
\IEEEdisplaynontitleabstractindextext
\IEEEpeerreviewmaketitle

\IEEEraisesectionheading{\section{Introduction} \label{sec:introduction}}
\IEEEPARstart{W}{ith} an estimated 100 billion neurons~\cite{hubel1979brain} and only 20~watts of power consumption~\cite{drubach2000brain}, the efficiency of a human brain is to this day unmatched by even the best AI accelerators. Spiking neural networks (SNNs) replicate the neural dynamics of biological brains more closely than conventional artificial neural networks (ANNs), therefore promising better efficiency~\cite{blouw2019benchmarking, wang2022efficient, gerlinghoff2022resource}. Information transfer between neurons in an SNN is achieved through {\em spike trains}, which are binary events with values of either one or zero. Each spiking neuron has a {\em membrane potential}. Every incoming spike will cause the membrane potential to be charged up by a certain amount according to the weight stored in the synapse. Once the membrane potential exceeds a predefined threshold, an output spike is fired to its successors. This describes the behavior of an integrate-and-fire (IF) neuron~\cite{koch1998methods}. Because input activations are binary, this type of neurons does not require multipliers. Instead, an \texttt{AND} operation is used to decide whether or not the membrane potential is modified. Other neuron models~\cite{hodgkin1952quantitative, izhikevich2003simple} might be more biologically accurate, but are computationally complex and hence not suitable for high-performance applications.

The IF neuron model can generally be implemented on general computing infrastructure, such as CPUs and GPUs. Because of their generality and the presence of floating-point arithmetic units, they usually achieve a high accuracy. At the same time, however, they contain hardware logic that is not directly needed for the computation and is hence detrimental to power and area efficiency. Application-specific integrated circuits (ASICs) can achieve high efficiency by only implementing the necessary neuron dynamics on hardware~\cite{nambiar20200}. Accordingly, it is important to consider the trade-off that exists between accuracy and performance, in light of the specific requirements of the application. Examples of ASICs are TrueNorth~\cite{esser2016convolutional}, SpiNNaker~\cite{painkras2013spinnaker}, and Loihi 1/2~\cite{davies2018loihi, intel2021taking}. Through user programming, a variety of neural models and learning rules can be deployed. But that also means that flexibility is limited to the capabilities of the existing hardware cores. With neuromorphic computing developing at a rapid pace, existing ASICs might not be able to keep up with new requirements. Field-programmable gate arrays (FPGAs), on the other hand, are reconfigured for every task and are greatly adapted to changes in network architecture, neuron model, etc.

Advances in packaging technology has led to FPGA devices which integrate multiple equivalent dies along a vertical stack. In Xilinx terminology, a die is referred to as {\em super logic region} (SLR). Vertical routing between SLRs is achieved via silicon interposers, which are special tiles with dedicated flip-flops. Signal crossings between SLRs are sparse when compared with the abundance of routing resources as part of the FPGA fabric within an SLR. In order to avoid congestion, is it imperative for scalable high-performance accelerators to consider individual SLRs for resource allocation, instead of viewing the FPGA device as a whole.

Mapping layers in an SNN to the SLRs in the device greatly influences the performance. Processing bottlenecks mainly arise in two places: (1) within an SLR if memory and compute resources are not carefully balanced, (2) between SLR if too many data need to be transferred over the scarce vertical interconnects. With DeepFire2 (DF2), we propose a hardware architecture which allows fine-grained control over how layers are allocated to SLRs. On one hand, splitting large layers across multiple SLRs, while considering data transfers, allows us to overcome resource limitation imposed by a single SLR. On the other hand, leaving complete SLRs unused for small networks allows them to be powered down, reducing the power consumption significantly.

As a result, DeepFire2 demonstrates great scalability across a wide range of network sizes. We achieved very high throughput, computing thousands of frames per second (FPS) even for large datasets, by almost optimally using FPGA resources and implementing a parallel datapath. We are also the first who use a neuromorphic hardware platform to accelerate an SNN classifying ImageNet data. This work builds upon the first version of DeepFire~\cite{aung2021deepfire} and adds novel techniques that considerably improve speed and scalability. The contributions of this paper can be summarized as follows.
\begin{itemize}
    \item A combination of {\em layer-wise} and the novel {\em split-kernel} mapping is proposed to distribute layers across multiple SLRs while keeping the resource usage balanced.
    \item A novel implementation of the logical \texttt{AND} operation of spiking neurons, which makes use of FPGA registers, to reduce the utilization of lookup tables for large networks.
    \item We achieved the highest clock frequency and throughput among all FPGA-based SNN implementations, thanks to the deep pipelining possible in DF2.
\end{itemize}

Spiking neural networks have been applied to large datasets in the past, but very long spike trains in the order of thousands were required~\cite{sengupta2019going, han2020rmp}. However, to harness the full potential of SNNs, short spike trains are preferred. This not only decreases the total number of operations needed to generate the classification result. It also reduces the network's latency~\cite{li2021free}. Therefore, we chose the minimal spike train length of one, while maintaining a competitive accuracy on many datasets.

The paper is organized as follows. In Section~\ref{sec:related}, we discuss various existing SNN implementations in ASIC and FPGA. Section~\ref{sec:architecture} introduces our DF2 hardware architecture. Deployment of various SNN networks on DF2 and results are presented in Section~\ref{sec:implementation}, followed by the conclusion in Section~\ref{sec:conclusion}.

\begin{figure*}[t]
    \centering
    \includegraphics[width=\textwidth]{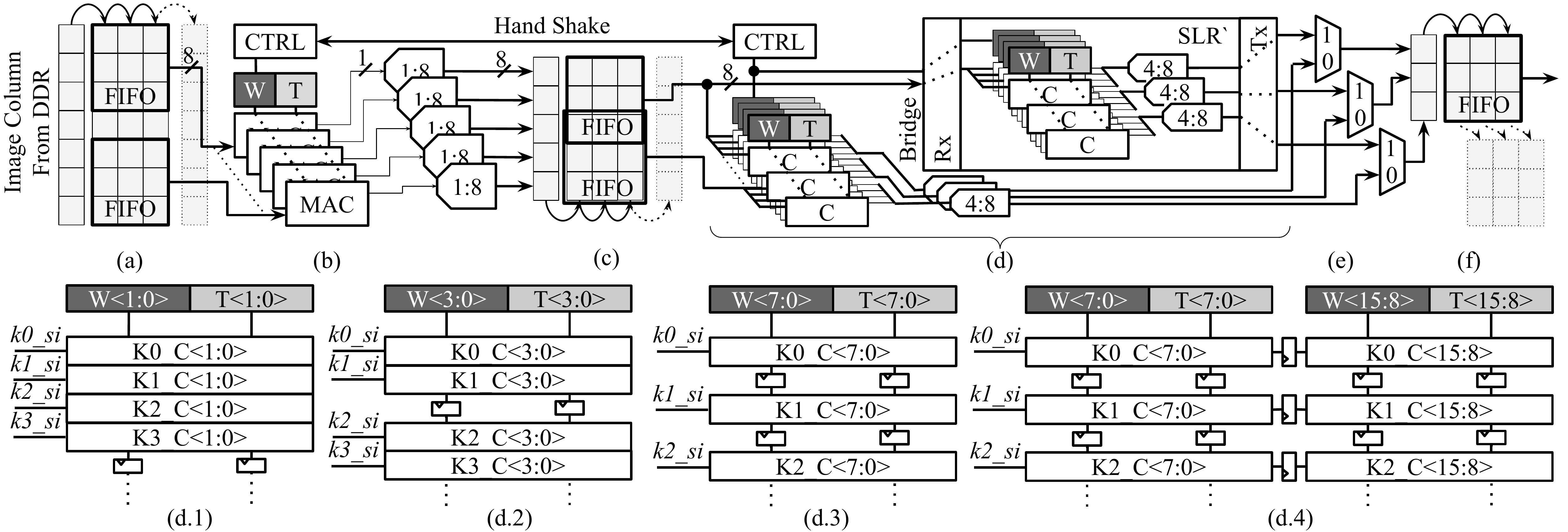}
    \caption{Architecture of DeepFire2's RTL. (a) two-stage image buffer for transduction layer, (b) transduction layer, (c) two-stage feature buffer (FBF) for convolution layer, (d) convolution layer with two-way kernel split, (d.1 - d.4) scaling of kernel arrays from 2 neuron cores per kernel to 16 cores per kernel, (e) MUX with round-robin scheduling to merge features from split-kernels, and (f) two-stage FBF for fully-connected layer.}
    \label{fig:df_nn}
\end{figure*}

\section{Related Work} \label{sec:related}
Traditional Von Neumann architectures lack the memory bandwidth and computing power necessary to achieve good energy efficiency and high throughput for neuromorphic workloads. Innovative chip architectures have been designed to overcome those bottlenecks~\cite{luo2021nc,yang2022coreset}. The SpiNNaker chip was created with multiple ARM cores for a parallel computation with shared on-chip synaptic memory storage to minimize the latency~\cite{painkras2013spinnaker}. TrueNorth~\cite{esser2016convolutional} and Loihi~\cite{davies2018loihi} remove the memory bottleneck entirely by dedicating an on-chip memory block for each computing neuron core, and the cores are connected through a Network-on-Chip (NoC). When thousands of cores are packed in a single chip, a data packet has to pass multiple nodes unnecessarily to reach the destination core. A hierarchical NoC was introduced in the Innatera chip to reduce the number of packet hops and its mixed-signal computing approach is promising to achieve low power~\cite{innatera}. There have also been attempts to use purely analog computations in neuromorphic hardware~\cite{valentian2019fully, luo2018fpga, shi2021fast}. Weight values are thereby written into resistive memory cells (RRAM) and input spikes switch the voltage over the resistors on and off. With this technology, whole vector-matrix multiplications can be implemented. But beside the noisy RRAM devices affecting the accuracy, the conversion between analog and digital values makes up a large fraction of the power consumption~\cite{shafiee2016isaac}.

Developing hardware is, however, only one part towards accelerating a spiking neural network. Compilers are needed, which map the neurons in the network to the resources in the chips. For Loihi, Intel developed the LCompiler, which optimizes the mapping between source and destination compartments for energy efficiency by using a greedy algorithm~\cite{lin2018mapping}. IBM proposed a new programming paradigm called {\em Corelets}, along with a programming language and library, for their neuromorphic chip TrueNorth. Corelets are neurosynaptic functions, which can be decomposed and mapped efficiently to the chip's neuron cores~\cite{amir2013cognitive}. General mapping algorithms, like MAMAP~\cite{zhang2020mamap}, partition SNNs in a way that minimizes the spike communication between cores. This positively influences throughput and energy consumption. Song~et~al.~\cite{song2021design} included buffer size in the trade-off. With an iterative clustering approach, they managed to improve those metrics compared to state-of-the-art. As can be seen, the fixed architecture of ASICs imposes various constraints on the network architecture and require mapping algorithms that increase the complexity deploying an SNN on neuromorphic hardware.

FPGAs, on the other hand, do not have a predetermined number of cores and, thus, overcome many architectural limitations. But a fixed number of overall compute resources, like lookup tables (LUTs), DSP slices and registers, requires area efficient hardware designs. Cluster computing is adopted in the Bluehive system for neuron simulation while overcoming the limited resources in each FPGA~\cite{moore2012bluehive}. Thanks to the off-chip memory, it can process up to 64k neurons per FPGA, similarly to Minitaur~\cite{neil2014minitaur}. When Thomas and Luk~\cite{thomas2009fpga} simulated the Izhikevich neuron model, 1000 neurons could be fitted into the on-chip memory of a single Virtex-5 FPGA. The neuron count increased to 1.44k when it was simulated with the more advanced Virtex-6~\cite{pani2017fpga}. In fact, the number of neurons can be a few orders of magnitude higher with appropriate quantization and when deploying it on modern FPGAs from AMD-Xilinx, which are now packed with 10$\times$ more memory capacity~\cite{xilinxDs890}. The aforementioned FPGA implementations are multilayer perceptrons (MLP) with fully-connected layers only. Their accuracy can, hence, be very limited when it comes to more complex image classification, such as Cifar-10. In recent years, FPGA designs for the acceleration of convolutional SNN have been published. Fang~et~al.~\cite{fang2020encoding} interpreted the spiking neurons as infinite impulse response filters. To process the data, a neural encoding scheme is suggested which converts continuous inputs into temporal spike trains. These can be efficiently implemented on the DSP slices in the FPGA. A performance analysis and optimization procedure automatically allocated resources to layers with higher latency. Ju~et~al.~\cite{ju2020fpga} reduced memory accesses of two-dimensional convolution and pooling layers by streaming data through shift registers. A two-dimensional array of convolution units then generates multiple output values in parallel. Additionally, time steps are executed in parallel through pipelining. Gerlinghoff~et~al.~\cite{gerlinghoff2021e3ne} built flexibility into their processing modules to enable reuse for feature maps of different sizes. High resource efficiency and low latency was thereby achieved through multiple levels of parallelism and use of an emerging spike encoding scheme. But reusing resources between layers and lack of DSPs limited their throughput. Lastly, Panchapakesan~et~al.~\cite{panchapakesan2021syncnn} proposed SyncNN, a synchronous design. To achieve very high throughput, input spikes over all time steps are aggregated and output spikes are generated in a single step, eliminating all but one forward pass per sample.

Nowadays, the model conversion from convolutional neural network to spike-based CNN had been studied extensively to achieve a reasonable accuracy~\cite{cao2015spiking, rueckauer2017conversion, shrestha2018slayer}. Most of those CNN to SNN conversion works, however, do not consider the target hardware. Our goal with DF2 is to compose a network architecture, which is most suited to the underlying SNN hardware. This enables optimal resource usage and pipeline balancing. Minimizing the spike train length leads to a reduction in latency and hardware complexity. The training of the network model is part of the compilation flow for DF2.

\section{DeepFire Hardware Architecture} \label{sec:architecture}
The DeepFire2 pipeline offers fine control over how spiking neural network layers are allocated to hardware resources in a multi-die FPGA device. This control is necessary to avoid performance bottlenecks which arise from resource scarceness and imbalance, which were the limiting factor of the {\em layer-wise} mapping of the first DeepFire (DF1)~\cite{aung2021deepfire}. Layer-wise mapping allocated network layers as a whole to super logic regions to avoid using large portions of the inter-SLR routing resources. No weights or partial sums have to be transferred, only the binary output spikes of the layers. But confining layers to only one SLR causes scalability issues.

Modern networks contain highly parameterized layers which cannot always be accommodated by the memory resources of a single SLR. Therefore, DeepFire2 introduces {\em split-kernel} mapping. which distributes kernels while keeping the signal crossings between SLRs to a minimum. By adjusting the splitting ratios, the designer can balance the resources of every SLR for optimal utilization. Split-kernel mapping comes at the cost of less than ten cycles of latency.

Figure~\ref{fig:df_nn} details the processing pipeline of the DF2 architecture. It generally alternates between neuron cores and feature buffers which will be described in Sections~\ref{sec:core} and~\ref{sec:fbf}, respectively. Alongside the neuron cores, there are controllers and memory units for weights and thresholds. The initial transduction layer converts integer inputs into spike trains.

\subsection{Split-Kernel Mapping} \label{sec:split_kernel}
Figure~\ref{fig:df_nn}~(d) shows the components used to compose a convolution layer. To precisely adjust the resource utilization and parallelism of the layer, its hardware modules are characterized in two dimensions:
\begin{enumerate}
    \item Number $\kappa$ of kernel units K per layer which compute the convolution of a $3\times3$ window in the feature map. This is determined by the feature map size.
    \item Number $\omega$ of parallel weight and threshold units W \& T which store weights and thresholds. This depends on the throughput requirements and available hardware resources.
\end{enumerate}

The arrangement of neuron cores C can be visualized as a two-dimensional array of size $\kappa\times\omega$. Each unit W \& T is associated with $\kappa$ neuron cores C. Using Figure~\ref{fig:df_nn} as an example, the feature map stored in (c) has 5 rows, implying $\kappa=3$ kernel units for (d). For real-world models, however, initial layers can have as much as 224 kernel units for ImageNet. In order to retain a high clock frequency and, hence, high throughput, re-timing registers are introduced to partition large neuron core arrays. In DF2, groups of 8 neuron cores are formed. The grouping for selected values of $\omega$ ranging from 2 to 16 is shown in Figure~\ref{fig:df_nn}~(d.1 to d.4). (d.1), where each kernel has 2 cores, 4 kernels (K$<$3:0$>$) are grouped. W$<$1:0$>$ and T$<$1:0$>$ signals are re-timed and passed to K$<$7:4$>$. On the other hand, 2 kernels are grouped when each kernel has 4 cores (d.2). When each kernel has more than 8 neuron cores, multiple groupings are formed within the kernel itself as shown in (d.3 and d.4). A group size of 8 allows good timing closure on our test platform, but the optimal number may vary for different platforms. Therefore, the DF2 IP can adapt by changing the grouping parameter.

\begin{figure}[t]
    \centerline{\includegraphics[width=\columnwidth]{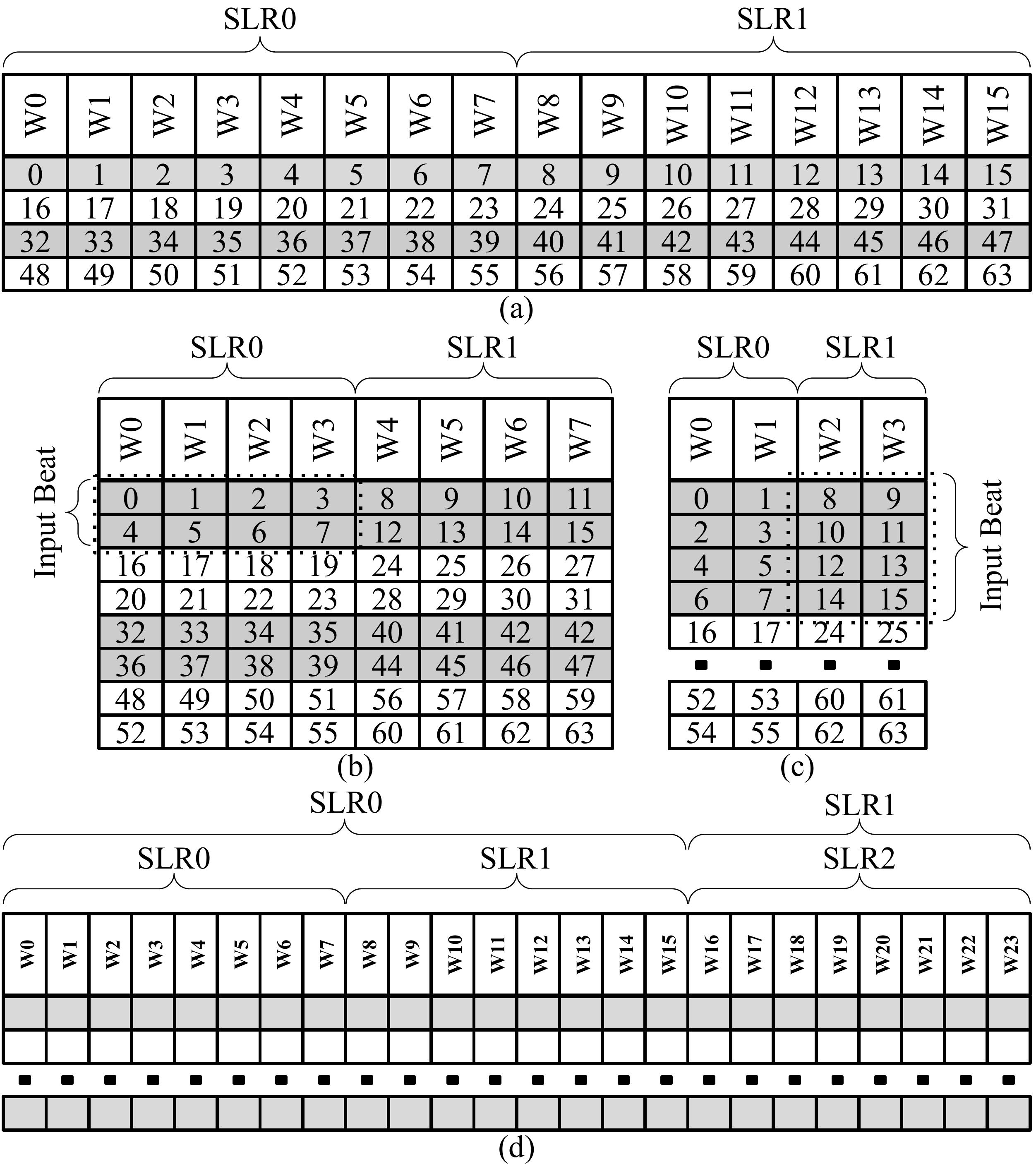}}
    \caption{Split-kernel mapping for 64 neurons across multiple SLRs and the numbers represent the neuron IDs. (a) 16 weight units and 16 neuron cores for 2 SLRs, (b) 8 weight units and 8 neuron cores for two-way split, (c) 4 weight units and 4 neuron cores for two-way split, (d) 24 weight units and neuron cores for either two-way or three-way splits. The input beats mark eight input spikes, which are delivered simultaneously over the \textit{si} bus.}
    \label{fig:df_mem}
\end{figure}

Split-kernel mapping allows the adjustment of the degree of parallelism to trade-off throughput and hardware resources. A larger $\omega$ instantiates more parallel neuron cores, increasing the utilization of DSPs and LUTs. In addition, $\omega$ influences the shape of memory blocks as demonstrated in Figure~\ref{fig:df_mem}~(a), (b) and (c). Splitting a layer across multiple SLRs along the $\omega$ dimension becomes necessary if one SLR does not have enough memory to accommodate all parameters. This applies especially to the last layers in a neural network.
Given the number of kernel weights (64 in this example), the user can choose different memory depths by cascading memory block primitives. The kernel map in Figure~\ref{fig:df_mem}~(a) uses shallower memory depth of 4 neurons with 16 weight units W0 to W15 to store the 64 kernel values, leading to 16 parallel cores. On the other end of the spectrum, it can also be reconfigured to 4 weight units with deeper memory and 4 cores to reduce the throughput and conserve computing resources (c). An {\em input beat} is the data transfer of input spikes, whose size is fixed to 8 according to the neuron core design (Section~\ref{sec:core}). How the eight spikes are mapped to the neuron cores is highlighted in the figure. Three-way splitting of the kernel across three SLRs is possible as shown in Figure~\ref{fig:df_mem}~(d), where one third of the weight units and cores are assigned to every SLR. Alternatively, the user can also choose an asymmetric two-way split, where two thirds of the kernel values are assigned to SLR0 and the remaining ones go to SLR1. 

When splitting layers across multiple SLRs, the primary SLR accommodates not only the allocated C, W \& T, but also the controller and the feature buffer. Secondary SLRs transceive only spiking features through RX and TX bridges. Since all the kernel operations are contained within each SLR, the strain on inter-SLR routing resources is relieved. The bridges are designed with multiple clock roots as suggested by~\cite{lavin2018rapidwright} for high-speed data crossing. The output features from the cores are merged at the MUXs in Figure~\ref{fig:df_nn}~(e), which are scheduled in round-robin style and route data to the feature buffer of the next layer. The the following sections, the pipeline components are explained in detail.

\subsection{Neuron Core Design} \label{sec:core}
\begin{figure}[t]
    \centerline{\includegraphics[width=\columnwidth]{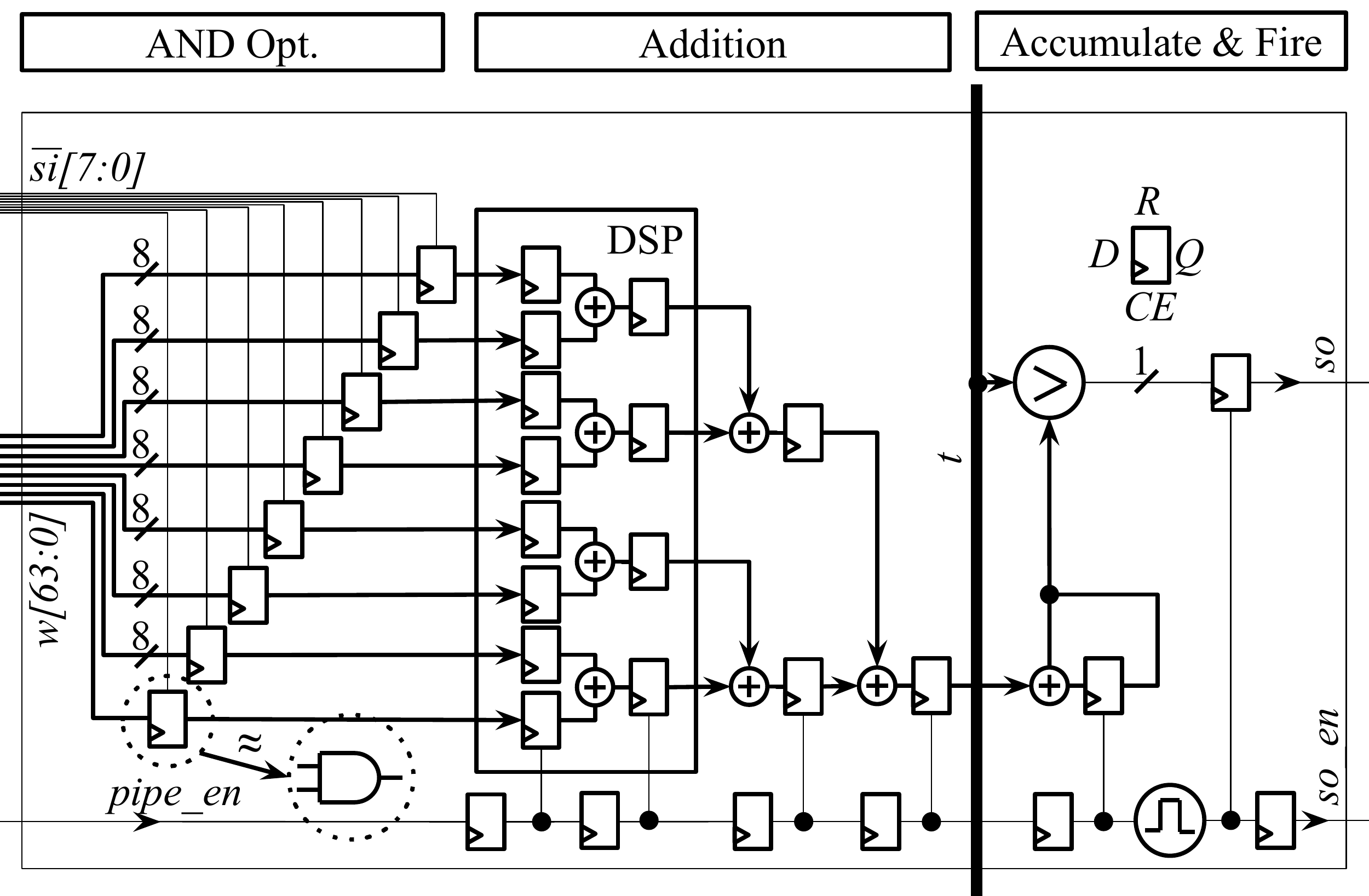}}
    \caption{Architecture of a single DeepFire2 neuron core, which is able to process eight binary input spikes~\textit{si} in parallel and generates spikes at neuron output~\textit{so}. It consists of registers as \texttt{AND} gates, an adder tree and an accumulation \& fire stage.}
    \label{fig:df_neuron}
\end{figure}

Our DF2 neuron core design shown in Figure~\ref{fig:df_neuron} is the RTL version of the integrate-and-fire (IF) neuron. It processes eight input spikes~\textit{si[7:0]} and their weights~\textit{w[63:0]} in parallel, where weights are quantized to 8~bits. The operation includes an \texttt{AND}, addition, and accumulate~\&~fire stage. When accelerating large networks, a substantial number of neuron cores are instantiated in the FPGA. This renders the previous DF1 neuron core design~\cite{aung2021deepfire} unusable as it consumes too many lookup table resources, limiting the scalability of the accelerator. Our improved neuron core design reduces the use of LUTs by routing weights~\textit{w} directly to the register input~D. The \texttt{AND} operation is carried out by connecting inverted input spikes~\textit{si} to the synchronous reset~R of the registers. If an input spike is not present the reset is asserted, which overrides input~D and sets the data output~Q to low in the next clock cycle.

Saving a few LUTs on a hardware module can have a great impact when it is instantiated many times to deploy large-scale models. Registers are also more abundant in Xilinx UltraScale+ devices, since each configurable logic block (CLB) contains eight 6-input LUTs, but 16 flip-flops. Moreover, using a 6-input LUT for an \texttt{AND} operation with only two inputs wastes resources. Lastly, these \texttt{AND} registers also act as the re-timing registers. They allow DF2 to be run at a higher clock speed, since they relax the placement distance between the result of the \texttt{AND} operation and the DSP slice of the adder tree. 

The adder tree consists of three stages, where the first stage is embedded in DSP blocks in SIMD configuration and the remaining two-stages are implemented in CLBs utilizing LUTs and carry-chains. The adder tree result is then accumulated and compared against the threshold parameter~\textit{t}, which was trained together with the weights. The neuron core fires an output spike, i.e. $\textit{so} = 1$, if the membrane potential exceeds~\textit{t}, and remains silent otherwise. The entire operation of the core is pipelined with an accompanying enable signal \textit{pipe\_en}, which streams through the core together with the data. That translates into a signal \textit{so\_en} at the output, which indicates a valid spike value.

\subsection{Feature Buffer} \label{sec:fbf}
As we adopt the data-flow architecture from FINN~\cite{umuroglu2017finn}, the feature buffer (FBF) is important to hold and transfer spike features seamlessly from one layer to another. In DF2, two-stage buffering of the feature maps is proposed as shown in Figure~\ref{fig:df_nn}~(a), (c) and (f). 

An intermediate buffer~(c) uses the first stage for feature column storage. It will gather all the output spikes for a particular column from the previous layer and pass them to the FIFO (first in, first out) buffer in the second stage. Each FIFO holds three feature columns for 3$\times$3 convolution. In the specific case of~(c), the feature map has five rows. Hence, three FIFOs are instantiated for the three overlapping kernel windows in vertical direction (depicted are two FIFOs). The spiking features are sent to the neuron core bus~\textit{si}. At the end of each convolution, the FIFOs will discard one column and, thus, shift the convolution windows in horizontal direction along the rows. Figure~\ref{fig:df_nn}~(a) and~(f) show special cases of the FBF, as they precede the transduction and fully-connected layers, respectively.

\subsection{Transduction Layer} \label{sec:trans_layer}
The transduction layer is the first layer in the DF2 network where the raw image data are transformed into spiking signals (Figure~\ref{fig:df_nn}~(b)). It closely follows the neuron core design described above. But since both weights~\textit{w} and activations are 8-bit operands, a multiplier is used instead of the \texttt{AND} operation. DSP slices are used to implement the multiplier. This is followed by the adder tree, accumulator and comparator. The decision of whether an output spike is fired or not, depends again on the membrane potential after accumulation and the trained threshold~\textit{t}. The binary spikes are queued in a buffer~\textit{(1:8)} to convert the data width from 1-bit spike values to 8-bit column features. The column features are forwarded to the next FBF.

\subsection{Controller} \label{sec:controller}
There is a controller at every SNN layer. It coordinates the kernel operation of the layer as well as performs handshake with the layer before and after. The controller at layer $n$ will only start the kernel operation upon meeting both of the following conditions.
\begin{enumerate}
  \item The second stage of the FBF in layer $n$ is \textbf{loaded} with the features required for the current convolutional window. 
  \item The first stage of the FBF in layer $n+1$ is \textbf{empty} that output column features can be stored when the kernel operation starts. 
\end{enumerate}

This condition ensures that there is no data overflow in the FBFs. When the first state of $\text{FBF}_{(n+1)}$ is full, it will naturally create a backpressure all the way back to the transduction layer until it is cleared. The backpressure scenario can be avoided by manipulating the number of the weight units $\omega$ in each layer (more details in Section~\ref{sec:network}).

\section{Implementation} \label{sec:implementation}
In this section, we discuss the DF2 implementation for various SNN architectures including the compilation, the network scalability and the DF2 sub-system. We evaluate the performance on a wide range of datasets and compare it to previous SNN implementations.

\subsection{Network Architecture} \label{sec:network}
\begin{table*}
    \caption{Implemented Network Structures}
    \begin{center}
    \begin{tabular}{*{5}{c}}
    \toprule
    \textbf{MNIST}
    &\textbf{Cifar-10}
    &\textbf{Cifar-100}
    &\textbf{Tiny-}
    &\textbf{ImageNet} \\
    &&& \textbf{ImageNet} & \\
    \addlinespace\hline\addlinespace
    pConv3-1-16~/~b1 &pConv3-1-64~/~b1  &pConv3-1-64~~/~b1 &pConv3-1-64~/~b1  &pConv3-1-64~/~b2  \\
    ~Conv2-2-16~/~b1 &pConv3-1-64~/~b1  &pConv3-1-112~/~b2 &~Conv2-2-64~/~b1  &~Conv2-2-64~/~b2  \\
                     &~Conv2-2-72~/~b1  &~Conv2-2-144~/~b2 &                  &                  \\
    \addlinespace\hline\addlinespace
    pConv3-1-32~/~b1 &pConv3-1-192~/~b2 &pConv3-1-192~/~b4 &pConv3-1-112~/~b2 &pConv3-1-112~/~b4 \\
    ~Conv2-2-32~/~b1 &pConv3-1-128~/~b2 &pConv3-1-216~/~b4 &~Conv2-2-144~/~b2 &~Conv2-2-144~/~b4 \\
                     &~Conv2-2-128~/~b2 &~Conv2-2-288~/~b4 &                  &                  \\
    \addlinespace\hline\addlinespace
    pConv3-1-64~/~b2 &pConv3-1-224~/~b4 &pConv3-1-480~/~u1 &pConv3-1-192~/~b4 &pConv3-1-200~/~b8 \\
    Conv3-2-64~/~b2  &pConv3-1-256~/~b4 &pConv3-1-504~/~u1 &pConv3-1-216~/~b4 &pConv3-1-288~/~b8 \\
                     &pConv3-1-224~/~b4 &pConv3-1-560~/~u1 &~Conv2-2-288~/~b4 &~Conv2-2-224~/~b8 \\
                     &~Conv2-2-288~/~b4 &~Conv2-2-560~/~u1 &                  &                  \\
    \addlinespace\hline\addlinespace
    Fc-128~/~u2      &Fc-560~/~u2       &Fc-1064~/~u2      &pConv3-1-480~/~b8 &pConv3-1-512~/~b16 \\
    Fc-10~/~b1       &Fc-10~/~b2        &Fc-100~/~u4       &pConv3-1-448~/~b8 &pConv3-1-448~/~b16 \\
                     &                  &                  &~Conv3-1-448~/~u1 &~Conv2-2-576~/~b16 \\
    \addlinespace\hline\addlinespace
                     &                  &                  &pConv3-1-512~/~u1 &pConv3-1-600~/~u2  \\
                     &                  &                  &pConv3-1-560~/~u1 &pConv3-1-576~/~u2  \\
                     &                  &                  &~Conv3-1-528~/~u1 &~Conv2-2-448~/~b16 \\
    \addlinespace\hline\addlinespace
                     &                  &                  &Fc-1200~/~u4      &Fc-1056~/~u8       \\
                     &                  &                  &Fc-200~/~u4       &Fc-1000~/~b8       \\
    \bottomrule
    \addlinespace
    \multicolumn{5}{l}{%
    \begin{minipage}{14cm}%
    \textbf{Conv3} = convolution with \textit{valid} padding, i.e. no padding; \textbf{pConv} = convolution with \textit{same} padding; \textbf{pConv3-1-64} = 3$\times$3 \textit{same} padding convolution with stride 1 and 64 output neurons; \textbf{Fc-1200} = fully-connected layer with 1200 neurons; \textbf{b2} = weight unit is 2$\times$ BRAM cascade; \textbf{u2} = weight unit is 2$\times$ URAM cascade.
    \end{minipage}}\\
    \end{tabular}
    \label{tab:network}
    \end{center}
\end{table*}

To characterize the DeepFire2 architecture for neural network of different sizes, we implemented models for a variety of datasets, as shown in Table~\ref{tab:network}. They closely follow VGG-9 and VGG-11 architectures with a customized neuron count for each layer. The neuron count is chosen to efficiently utilize the fixed-size BRAM and URAM blocks in the device. For example, mapping a kernel with 112 neurons to 4kB large memory results in instantiation of 8 BRAM blocks, each containing the weights of 14 neurons. This leads to 88.8\% utilization for each 4kB-BRAM. 128 neurons, as per original VGG configuration, results in 16 BRAMs instantiation with 8 neurons per BRAM block (50.8\% utilization). A number between 8 and 16 is not possible because of the byte write operation to the FBF. In summary, the number of weight units \textit{W} for weight parameters in each kernel must satisfy the following condition.

\begin{equation} \label{eqn:w}
  W =
    \begin{cases}
      2^{i} & \text{if $0\le i\le 2$}\\
      2^{3}(i-2) & \text{if $i>2$}
    \end{cases}
\end{equation}

where $i$ is the integer greater or equal to zero. The maximum pooling layer in the conventional VGG architecture is replaced by a 2$\times$2 convolution with a stride of 2 and \textit{valid} padding. Hence, DF2 pooling layers carry trainable weights. Table~\ref{tab:network} also presents the cascading parameters for BRAM and URAM for each layer.

The neural network architecture dictates the allocation of memory blocks. The default BRAM and URAM modules in the FPGA can be cascaded to form larger memory. As layers become wider, more parameters are used and, hence, deeper memory is required to store these parameters. Each URAM block has a capacity 8$\times$ as large as a BRAM block. Moreover, URAM has a deep internal pipeline for data readout and thus, it is capable of cascading multiple URAM blocks without sacrificing its maximum clock frequency Fmax~\cite{aschule2021achieving}. Whereas for BRAM, its Fmax performance drops when blocks are cascaded. Therefore, the usage of URAM is preferred for wider layers later in the model. For our larger models, however, up to 16 BRAM blocks had to be cascaded due to a lack of URAM resources, which lead to a slight drop in frequency Fmax.

The memory depth influences parameters like the number of weight units $\omega$. Making memory blocks shallower, but wider, increases parallelism and resources. The initial layers, where high-resolution feature maps get processed, are instantiated with shallow memory depth and many neuron cores. The deeper layers with lower feature resolution are allocated deeper memory and a lower number of cores. In an optimal setting, the data flow from the first layer to the last layer will be seamless without any backpressure in the pipeline. Through this setting, the DF2 IP can be adapted to various FPGA platforms and it is able to maximize the utilization of compute resources for maximum performance and low power consumption.

\subsection{Scalability} \label{sec:scale}
\begin{figure}[t]
    \centerline{\includegraphics[width=\columnwidth]{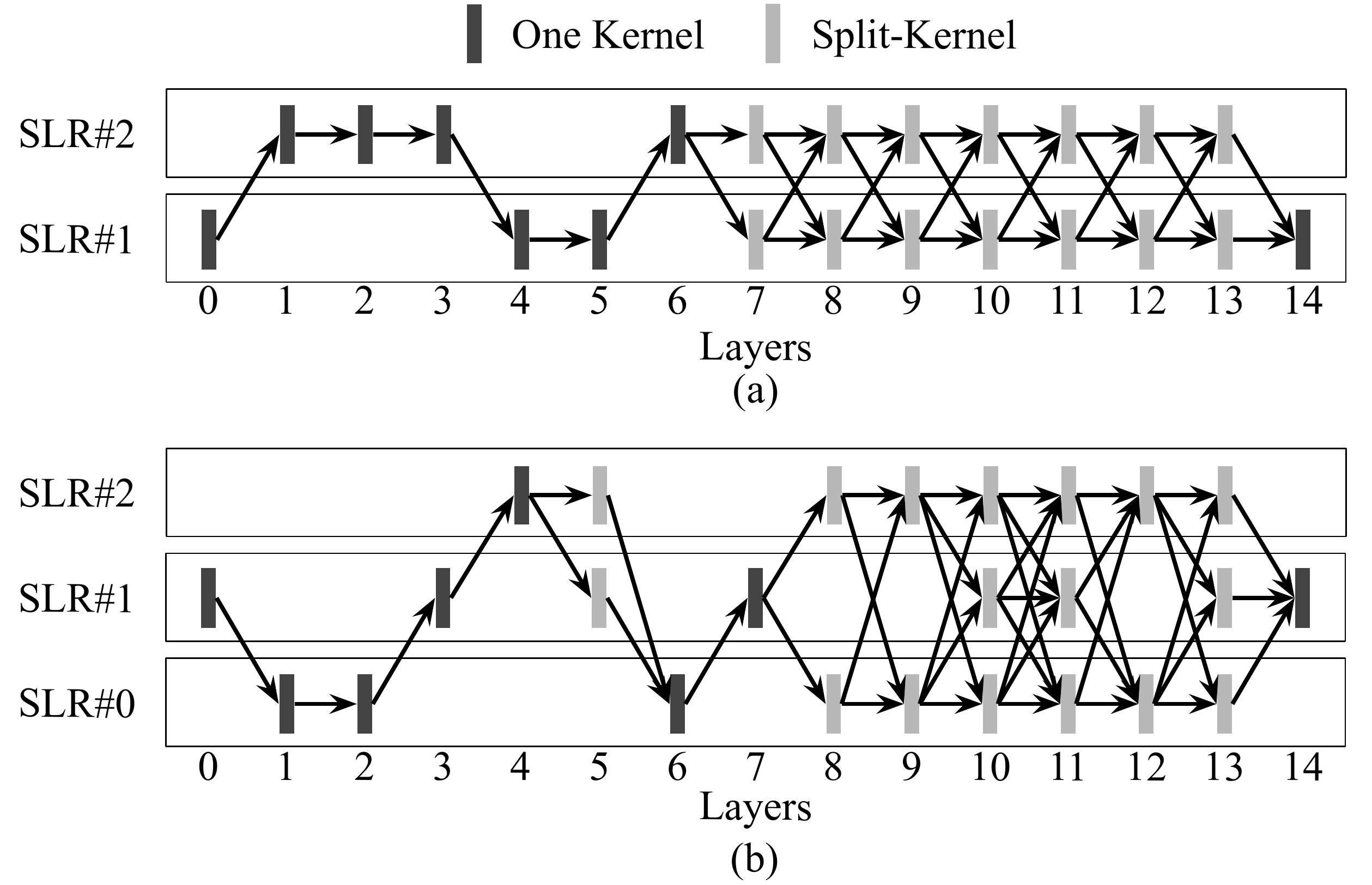}}
    \caption{The dataflow chart of VGG-11 mapping on multiple SLRs. (a) Tiny-ImageNet, and (b) ImageNet.}
    \label{fig:df_dataflow}
\end{figure}

\begin{figure*}[htbp]
    \centering
    \includegraphics[width=\textwidth]{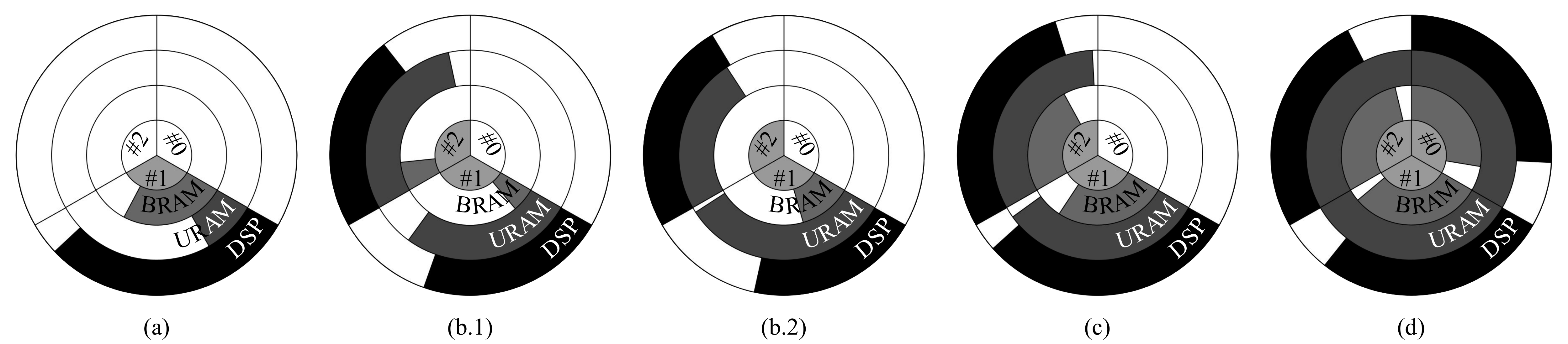}
    \caption{Memory and DSP utilization of DF2 in various SNN architectures. Each pie chart is equally split into three, each representing one SLR and its resource utilization in percentage. (a) Cifar-10 on SLR\#1, (b.1) Cifar-100 on SLR\#1/2 with both layer-wise and split-kernel mapping, (b.2) Cifar-199 on SLR\#1/2 with layer-wise mapping only (c) Tiny-ImageNet on SLR\#1/2, and (d) ImageNet on SLR\#0/1/2.}
    \label{fig:df_config}
\end{figure*}

\begin{figure}[t]
    \centerline{\includegraphics[width=\columnwidth]{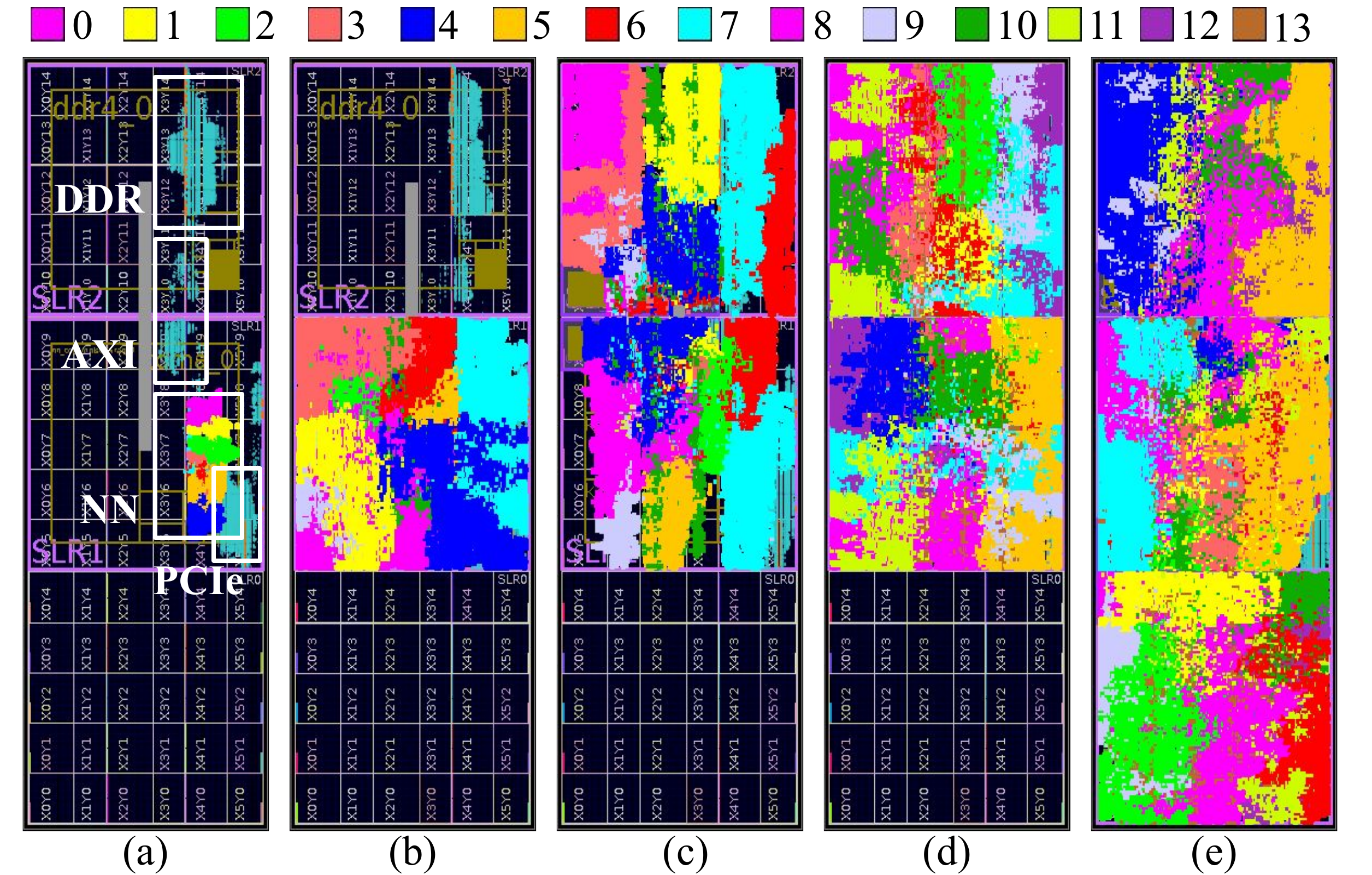}}
    \caption{The FPGA post implementation mapping of SNN on VU9P 3$\times$ SLRs. (a) MNIST, (b) Cifar-10, (c) Cifar-100, (d) Tiny-ImageNet, and (e) ImageNet.}
    \label{fig:df_fpga}
\end{figure}

The first version of DeepFire~\cite{aung2021deepfire} demonstrated its scalability by {\em layer-wise} SLR mapping. This allows it to scale for a deeper network. The limitation of this strategy is that layers can no longer be mapped into a single SLR, if they have too many parameters. As an example, the deeper layers Fc-1200 and Fc-1056 from Table~\ref{tab:network} require more memory resources than a single SLR can offer and it would not be possible to map these layers using the original DF1. DF2 introduces the split-kernel mapping for those heavily parameterized layers. Therefore, DF2 can support not only deeper but also wider networks.

The implemented VGG-11 mappings of Tiny-ImageNet and ImageNet are shown in Figure~\ref{fig:df_dataflow} with respect to their dataflow. A few initial layers with less weight parameters adopt layer-wise mapping and the following heavily parameterized layers adapt split-kernel mapping. DF2 provides the flexibility to carry out optimizations on a higher level of abstraction, which is not within the scope of this paper. While we mapped layers manually by estimating their required numbers of SLRs, other more optimal mapping strategies could certainly be applied. The ultimate goal of this mapping is 
\begin{itemize}
    \item to enable efficient distribution of FPGA resources across multiple SLRs and avoid bottlenecks caused by lack of memory or compute resources,
    \item to maximize the use of available resources in line with overall performance goals.
\end{itemize}

The results of the Vivado post-implementation on five different SNN architectures are shown in the Figure~\ref{fig:df_fpga}. Each unique color represents a layer in the network. As the network grows from Figure~\ref{fig:df_fpga}~(a) to (e), DF2 scales from one SLR to three SLRs. In theory, it can scale to an unlimited number of SLRs but in practice, due to the limited scale of production, only up to four SLRs per package are available commercially~\cite{xilinxDs890}.

Figure~\ref{fig:df_config} shows the distribution of DF2 resources with different SNN architectures. When an SLR is assigned with the compute intensive initial layers of the network, it will consume a lot of DSP resources but yet, a lot of on-chip memory will be left unused. On the other hand, the deeper layers will deprive the SLR of all the on-chip memory resources and leave many of the DSP resources untouched. Regardless of the number of assigned SLRs, the flexible DF2 mapping is capable of distributing the resource utilization for both memory and DSP evenly across multiple SLRs. That prevents a single type of resource from capping the performance. An additional advantage is a good power distribution throughout the large silicon substrates. Hence, the system is less prone to local drops in supply voltage, making it more robust. Since all the assigned SLRs share the resources almost equally, any localized routing congestion is also minimized. Figure~\ref{fig:df_config}~(a) shows the resource utilization of Cifar-10 and due to its small memory footprint, its implementation is confined to SLR1. Cifar-100 is implemented in SLR1 and SLR2 and (b.1) shows the equal distribution of resources among the two SLRs when both layer-wise and split-kernel mappings are applied. The resources are unbalanced in two SLRs of (b.2) despite putting the best effort in layer-wise only mapping. In that case, the URAM resources in SLR1 are exhausted, limiting further performance improvements. DF2 is also able to map the larger Tiny-ImageNet and ImageNet network in two and three SLRs efficiently, achieving almost 80\% utilization in both memory and compute resources ((c) and~(d)).

\subsection{System-Level Implementation} \label{sec:system}
The DF2 subsystem consists of PCIe DMA, DDR memory, AXI DMA, AXI interconnect, and a clock generator. PCIe DMA is mainly used for the host PC to send images and network parameters (weights and thresholds) to the FPGA DDR, and then reads back the classification results from the same DDR. AXI DMA is responsible for sending images from DDR to DF2 and writing classification results from DF2 to DDR. AXI interconnects provide routing for all the AXI-related IPs and enable clock domain crossings. The clock generator uses the reference clock from PCIe to generate a higher frequency clock for DF2 IPs. The post-placement result of the subsystem is highlighted in Figure~\ref{fig:df_fpga}~(a). 

\subsection{Compilation Flow} \label{sec:compile}
\begin{figure}[t]
    \centerline{\includegraphics[width=\columnwidth]{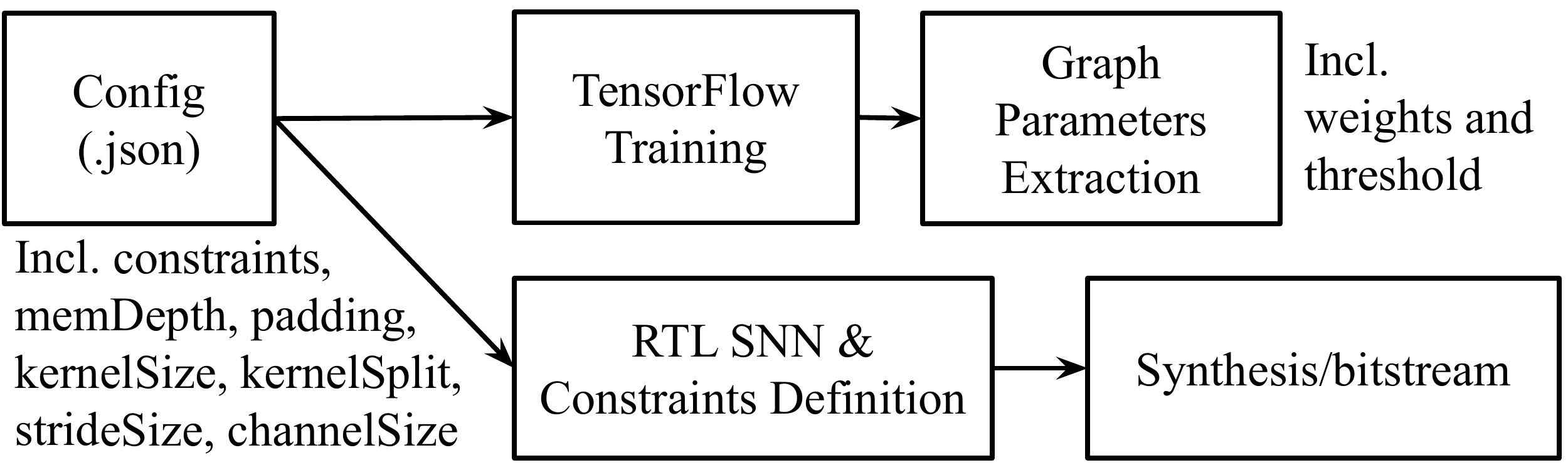}}
    \caption{Compilation flow for DF2 framework.}
    \label{fig:df_flow}
\end{figure}

The DF2 compilation flow (Figure~\ref{fig:df_flow}) starts with the user defined \texttt{config.json} file. It contains SNN network information for TensorFlow training such as kernel size, channel and stride information for each layer, etc. The SNN training is performed just like CNN training using standard back-propagation with batch normalization at every layer. This version of DeepFire supports padding which positively affects the classification accuracy. We start training with a leaky ReLU activation to avoid overfitting. Later in the training, we switch to \textit{tanh} activation. We also make use of data augmentation. The outputs of the activation function are binarized and a straight-through estimator is applied for back-propagation. Finetuning the model with spike activations helps to maintain a high accuracy while using fewer neurons.

FPGA-specific mapping information, such as kernel split and memory depth for each layer, are used to extract the weights and threshold parameters from the TensorFlow graph in accordance with the data sequence the FPGA is expecting to receive. During the extraction, 8-bit quantization is performed. The threshold value of each neuron is computed by folding the activation with the batch normalization parameters and quantizing them with the respective scale factor from the weight quantization. The same \texttt{config.json} is also used to generate the RTL wrapper and FPGA-specific constraints for the SLR bridges and the floor planning. We created a tcl script in the Vivado tool to build the DF2 subsystem from the DeepFire IPs. Once the subsystem is built, it will automatically trigger the synthesis and finally, produce an FPGA bitstream.

\subsection{Performance Comparison} \label{sec:performance}
\begin{table*}[htbp]
    \caption{DeepFire (DF2) Hardware Performance Benchmark with Prior Works}
    \begin{center}
    \setlength{\tabcolsep}{3.9pt}
    \begin{tabular}{l *{3}{c} | *{6}{r} | *{5}{r}}
        
        \toprule
        
        \multicolumn{4}{c|}{} & \multicolumn{6}{c|}{\textbf{Performance}} & \multicolumn{5}{c}{\textbf{Resources}} \\
        \textbf{Dataset} & \textbf{Ref.} &\textbf{Param.} & \textbf{Platform} & \textbf{MHz} & \textbf{Acc\%} & \textbf{kFPS} & \textbf{GOPS} & \textbf{kFPS/W} & \textbf{GOPS/W} & \textbf{\#SLR} & \textbf{kLUT} & \textbf{BRAM} & \textbf{URAM} & \textbf{DSP} \\
        
        \addlinespace\cline{1-15}\addlinespace
        
        \multirow{10}{*}{MNIST}
        & \cite{gerlinghoff2021e3ne}      & 413k & VU13P  & 200 & 99.3  & 2.45 & 2.07   & 0.68 & 0.58   & \multirow{5}{*}{1} & 41  & -    & 0   & 0    \\
        & \cite{fang2020encoding}         & 413k & ZCU102 & 125 & 99.2  & 2.12 & 6.72   & 0.47 & 1.49   &                    & 156 & 282  & 0   & 1794 \\
        & \cite{zhang2019asynchronous}    & 1.1M & VC707  & -   & 98.1  & 0.9  & 0.36   & 1.29 & 0.52   &                    & -   & -    & 0   & 0    \\
        & \cite{ju2020fpga}               & 252k & ZCU102 & 150 & 99.14 & 0.16 & 2.45   & 0.04 & 0.61   &                    & 125 & 264.5& 0   & 0    \\
        & \cite{panchapakesan2021syncnn}  & 230k & ZCU102 & 200 & 99.3  & 13.1 & 41.55  & -    & -      &                    & 48  & -    & -   & 251  \\
        & DF1\cite{aung2021deepfire}      & 252k & ZCU102 & 500 & 99.14 & 40.1 & 1.07e3 & 5.64 & 1.51e2 &                    & 55  & 138.5& 0   & 271  \\
        & \textbf{DF2} & \textbf{140k} & \textbf{VCU118} & \textbf{600} & \textbf{99.2} & \textbf{79} & \textbf{5.43e2} & \textbf{10.26} & \textbf{70.54} & & \textbf{8.8} & \textbf{30} & \textbf{4} & \textbf{132} \\
        
        \addlinespace\cline{2-15}\addlinespace
        
        & \cite{fang2020encoding}         & - & RTX5000   & 1620 & 99.2 & 0.864 & 2.74      & 0.014  & 4.44e{-2} & - & - & - & - & - \\
        & \cite{ju2020fpga}               & - & i7-6700K  & 4000 & 98.9 & 0.004 & 6.12e{-2} & 7e{-5} & 1.07e{-3} & - & - & - & - & - \\
        & \cite{esser2015backpropagation} & - & TrueNorth & -    & 99.4 & 1.00  & 1.97      & 12.95  & 25.46     & - & - & - & - & - \\
        & \cite{fang2020encoding}         & - & Loihi     & -    & 94.7 & 0.097 & 0.31      & 0.4    & 1.27      & - & - & - & - & - \\
        
        \addlinespace\cline{1-15}\addlinespace
        
        \multirow{4}{*}{Cifar-10}
        & \cite{gerlinghoff2021e3ne}      & -   & VU13P  & 150 & 80.6 & 0.043 & 5.40   & 0.01 & 1.26   & 1 & 48  & -   & 0   & 0    \\
        & \cite{panchapakesan2021syncnn}  & 12M & ZCU102 & 200 & 90.8 & 0.062 & 28.34  & -    & -      & 1 & -   & -   & -   & -    \\
        & DF1\cite{aung2021deepfire}      & 12M & VCU118 & 425 & 81.8 & 28.3  & 1.21e4 & 0.99 & 4.23e2 & 3 & 386 & 969 & 385 & 2963 \\
        & \textbf{DF2} & \textbf{4.6M} & \textbf{VCU118} & \textbf{550} & \textbf{87.1} & \textbf{23} & \textbf{1.04e4} & \textbf{1.15} & \textbf{5.18e2} & 1 & \textbf{125} & \textbf{511} & \textbf{80} & \textbf{2025} \\
        
        \addlinespace\cline{2-15}\addlinespace
        
        & \cite{esser2016convolutional} & - & TrueNorth & - & 89.3 & 1.25 & 2.58e3 & 0.76 & 1.58e3 & - & - & - & - & - \\
        
        \addlinespace\cline{1-15}\addlinespace
        
        \multirow{4}{*}{Cifar-100}
        & \cite{gerlinghoff2021e3ne} & - & VU13P  & 150 & 65.0 & 0.006 & 91.33 & 0.001 & 18.27 & 1 & 88 & - & 0 & 0 \\
        & \textbf{DF2} & \textbf{17.8M} & \textbf{VCU118} & \textbf{500} & \textbf{65.9} & \textbf{11.6} & \textbf{1.55e4} & \textbf{0.389} & \textbf{5.21e2} & \textbf{2} & \textbf{183} & \textbf{289} & \textbf{452} & \textbf{2881} \\
        
        \addlinespace\cline{2-15}\addlinespace
        
        & \cite{esser2016convolutional} & - & TrueNorth & - & 65.5 & 1.53 & 3.16e3 & 0.92 & 1.89e3 & - & - & - & - & - \\
        
        \addlinespace\cline{1-15}\addlinespace
        
        Tiny-Imagenet
        & \textbf{DF2} & \textbf{23.6M} & \textbf{VCU118} & \textbf{450} & \textbf{46.7} & \textbf{10.3} & \textbf{1.40e4} & \textbf{0.270} & \textbf{3.67e2} & \textbf{2} & \textbf{251} & \textbf{1041} & \textbf{616} & \textbf{4002} \\
        
        \addlinespace\cline{1-15}\addlinespace
        
        ImageNet     
        & \textbf{DF2} & \textbf{36.8M} & \textbf{VCU118} & \textbf{450} & \textbf{40.1} & \textbf{1.56} & \textbf{2.11e4} & \textbf{0.033} & \textbf{4.47e2} & \textbf{3} & \textbf{371} & \textbf{1757} & \textbf{960} & \textbf{5400} \\
        
        \bottomrule
    \end{tabular}
    \label{tab:performance}
    \end{center}
\end{table*}

\begin{figure}[t]
    \centerline{\includegraphics[width=\columnwidth]{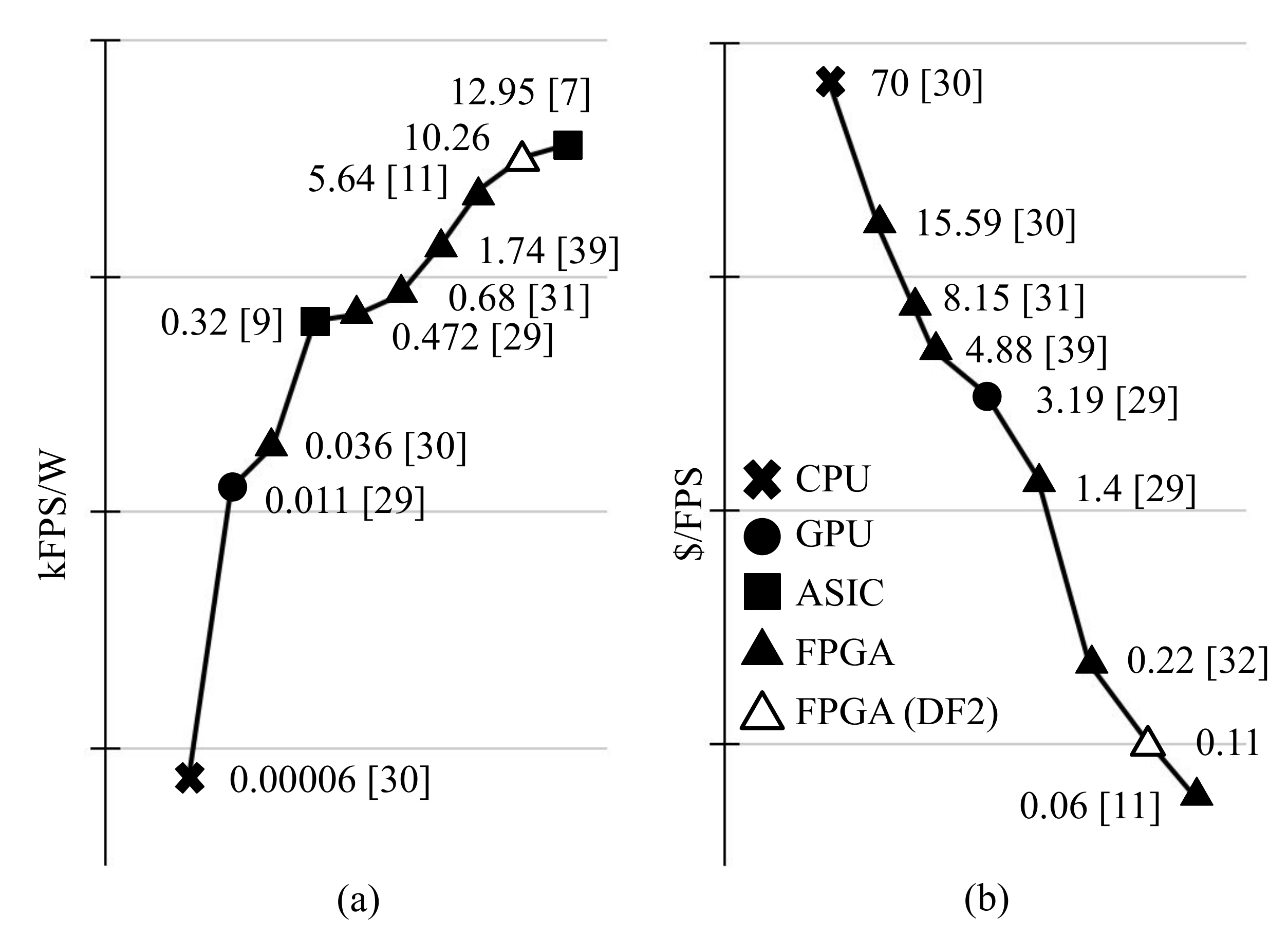}}
    \caption{(a) MNIST image classification kFPS/watt across different hardware implementation. kFPS/Watt is adjusted based on the assumption that every halving of process node leads to 40\% improvement. (b) cost-effectiveness (\$/FPS) of the MNIST implementation across various platforms.}
    \label{fig:df_pwr}
\end{figure}

A detailed performance comparison between DeepFire2 and previous SNN hardware implementations is provided in Table~\ref{tab:performance}. The table rows are grouped by dataset. While there have been plenty implementations for smaller datasets, like MNIST and Cifar-10, only DF2 is able to handle Tiny-ImageNet and ImageNet. In MNIST image classification, DF2 produces 10.26~kFPS/W which is only 20\% shy of TrueNorth's ASIC efficiency (see also Figure~\ref{fig:df_pwr}~(a)). At the same time, DF2 is at least one order of magnitude more energy efficient than other FPGA implementations~\cite{ju2020fpga, zhang2019asynchronous, fang2020encoding, gerlinghoff2021e3ne, aung2021deepfire}, and three and five orders of magnitude superior to GPU and CPU, respectively~\cite{ju2020fpga, fang2020encoding}. The maximum throughput of DF2 is 79~kFPS which almost doubles the original DF1 design~\cite{aung2021deepfire}. This is due to the improved pipelining at the neuron core level and the higher DF2 clock rate. DF2 throughput for MNIST can be much higher if we reduce the depth of the weight units in the network and deploy more neuron cores. DF2 Fmax is 600~MHz which is at least 3$\times$ higher than non-DF FPGA implementations. ASICs such as TrueNorth can achieve a relatively high throughput at 1~kFPS. However, it falls short of our FPGA throughput. GPUs acceleration is much superior to CPU throughput in general, but their energy efficiency is limited by the dataflow being bound to external DDR memory.

In addition to the frames-per-second, we report the performance of the listed neuromorphic architectures in GOPS (giga-operations per second) as a method of normalizing the throughput metric, thus making it independent of the processing required by a specific SNN model. Our GOPS measurements are calculated as the product of the ANN-equivalent number of operations per inference and the number of inferences per second. Similarly, we evaluate the power efficiency in GOPS per Watt.

Cost-effectiveness is an important consideration when it comes to large-scale deployments, for example in data centers. It is determined by normalizing the cost of the board to the throughput in FPS. How effectively the resources of a given device are used, i.e. the choice of compute architecture and various optimization strategies, greatly influences cost-effectiveness. In that regard, CPU is the least cost-effective solution and FPGA-based DF1~\cite{aung2021deepfire} and DF2 are the most cost-effective implementations across all hardware platforms. Figure~\ref{fig:df_pwr}~(b) also shows examples of FPGA implementations being far worse than the GPU counterpart. The cost-effectiveness of TrueNorth and Loihi cannot be measured because they are not commercially available.

When it comes to Cifar-10 classification, DF2 is still the most energy efficient SNN on an FPGA. Since we trade off the compute resources for better compute efficiency, our throughput is slightly less than~\cite{aung2021deepfire}. Better throughput and energy efficiency is expected in DF2 than that of~\cite{gerlinghoff2021e3ne} since we store all the weight and activation values on-chip. Furthermore, DF2 spike trains are shorter, while not sacrificing or even improving the classification accuracy. TrueNorth is 23\% better than DF2 in terms of FPS/W despite having to bridge four chips to map the large SNN. For the Cifar-100 dataset, TrueNorth has to bridge eight chips to accommodate a larger SNN and it is almost on par with DF2's energy efficiency, which bridges only two SLRs. 

DF2 demonstrates the mapping capability of a larger SNN for Tiny-ImageNet. DF2 achieves 46.7\% accuracy while its mapping is limited to two SLRs. Despite using over 90\% of on-chip memory and DSPs across two SLRs, DF2 can still clock at 450~MHz delivering 10.3~kFPS. DF2 also shows its scalability across multiple SLRs without compromising its Fmax for an ImageNet SNN, where it utilizes all three SLRs depleting almost all the on-chip DSPs and the memory resources (Figure~\ref{fig:df_config}~(d)). Even when operating at the limits of the VU9P FPGA, it can maintain a high performance. But approaching those limits does not allow the storage of all of the VGG-11 parameters, and the ImageNet accuracy on DF2 is limited to 40\%. This shows the general trade-off between accuracy and performance when neuromorphic hardware is compared to software implementations. But avoiding DDR accesses for weight transfer has significant advantages in terms of power consumption. The DF2 ImageNet performance is 21~TOPS (tera-operations per second) which about 78\% of the total DSP performance and it delivers approximately 86 images per TOPS.

Overall, DF2 is one of the most energy efficient and most scalable SNN implementations on FPGA, and it can provide the highest throughput among peers.

\section{Conclusion} \label{sec:conclusion}
DeepFire2 shows significant improvements over our first version of DeepFire~\cite{aung2021deepfire}. The addition of split-kernel mapping improved the balancing of computations between SLRs and helped deepen the pipeline. We maximize the DSP usage to achieve a high FPS/TOPS ratio. A high clock frequency of over 450~MHz is enabled by uniformly utilizing all hardware resources in the device (DSP, LUTs, memory). This is also enabled by avoiding excessive LUT-usage for \texttt{AND} operations and using available registers instead. Constraining smaller networks to only utilize one or two SLRs also improves power efficiency. Moreover, DF2 is the first neuromorphic IP in literature that can deploy a full-scale ImageNet implementation on an actual FPGA.

\ifCLASSOPTIONcompsoc
  \section*{Acknowledgments}
\else
  \section*{Acknowledgment}
\fi

This work was supported by the Singapore Government’s Research, Innovation and Enterprise 2020 Plan (Advanced Manufacturing and Engineering domain) under Grant A1687b0033.

\ifCLASSOPTIONcaptionsoff
  \newpage
\fi

\bibliographystyle{bib/IEEEtran}
\bibliography{bib/references}

\begin{IEEEbiography}[{\includegraphics[width=1in,height=1.25in,clip,keepaspectratio]{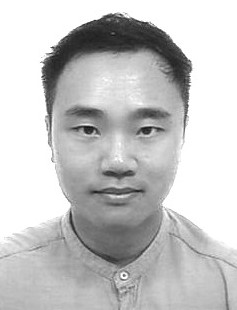}}]{Myat Thu Linn Aung}
    received the B.Eng. and Ph.D. degrees in electrical and electronic engineering from Nanyang Technological University (NTU), Singapore, in 2010 and 2016, respectively. From 2010 to 2011, he was with VIRTUS, IC Design Centre of Excellence, NTU, where he was involved in the development of motion-detection image sensors. He joined Xilinx Asia Pacific as a product quality engineer in 2016. In 2020, he was with the Institute of High Performance Computing in Singapore. His current research involves FPGA-based acceleration for neuromorphic computing.
\end{IEEEbiography}

\begin{IEEEbiography}[{\includegraphics[width=1in,height=1.25in,clip,keepaspectratio]{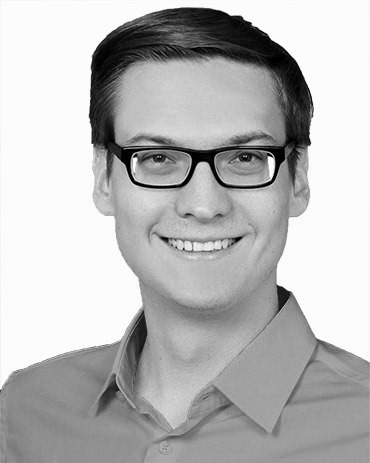}}]{Daniel Gerlinghoff}
    was awarded a master’s degree in integrated circuit design at Nanyang Technological University, Singapore. As part of his dissertation, he implemented a neural network inference accelerator on FPGA, which was tightly constrained by power and logic resources. After his graduation in 2020, he continues research on FPGA-based machine learning and its applications as a research engineer at Institute of High Performance Computing, Agency for Science Technology and Research in Singapore.
\end{IEEEbiography}

\begin{IEEEbiography}[{\includegraphics[width=1in,height=1.25in,clip,keepaspectratio]{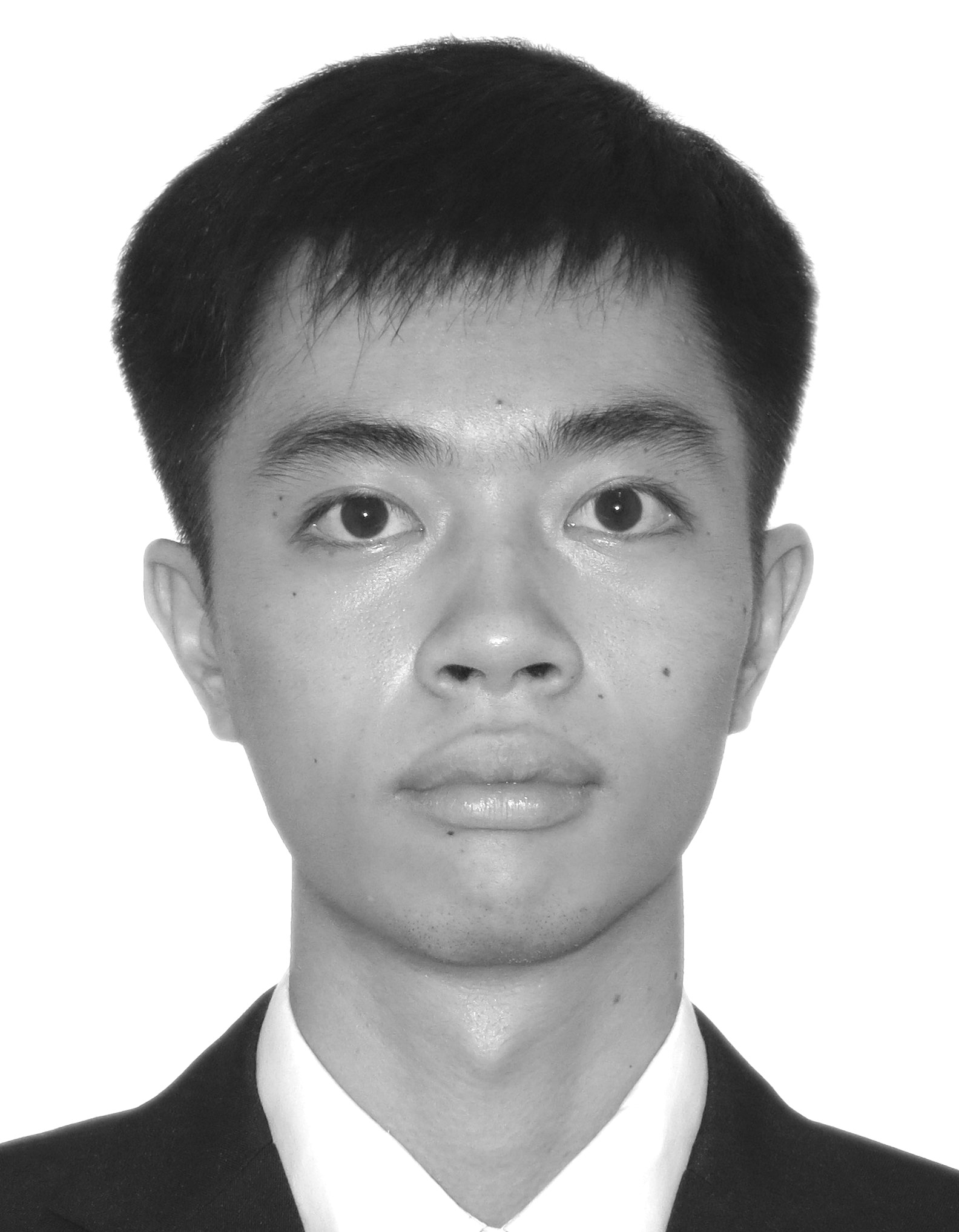}}]{Chuping Qu}
    received the B.Eng. degree (Hons.) in electrical engineering from the National University of Singapore, Singapore, in 2017. He did his university internship with Xilinx Asia Pacific Pte. Ltd. for the characterization of memory devices in FPGA. He is currently a Research Engineer with the Institute of High Performance Computing, Agency for Science, Technology and Research, Singapore. His current research interests include neuromorphic computing, hardware implementation on field-programmable gate array and artificial intelligence.
\end{IEEEbiography}

\begin{IEEEbiography}[{\includegraphics[width=1in,height=1.25in,clip,keepaspectratio]{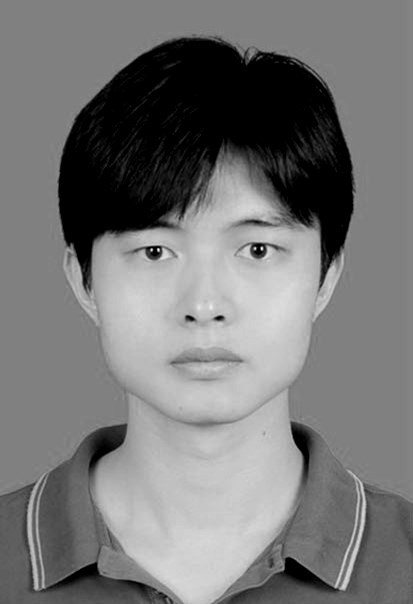}}]{Liwei Yang}
    is currently a research scientist at Institute of High Performance Computing (IHPC), Agency for Science, Technology and Research (A*STAR), Singapore. His research interests include high-performance computing, reconfigurable computing, neuromorphic computing, optimization in EDA toolchain and compilation for efficient AI and deep learning applications. He received his Ph.D. degree in computer science from Nanyang Technological University (NTU), and both M.S. and B.S. degrees from University of Electronic Science and Technology of China (UESTC).
\end{IEEEbiography}

\begin{IEEEbiography}[{\includegraphics[width=1in,height=1.25in,clip,keepaspectratio]{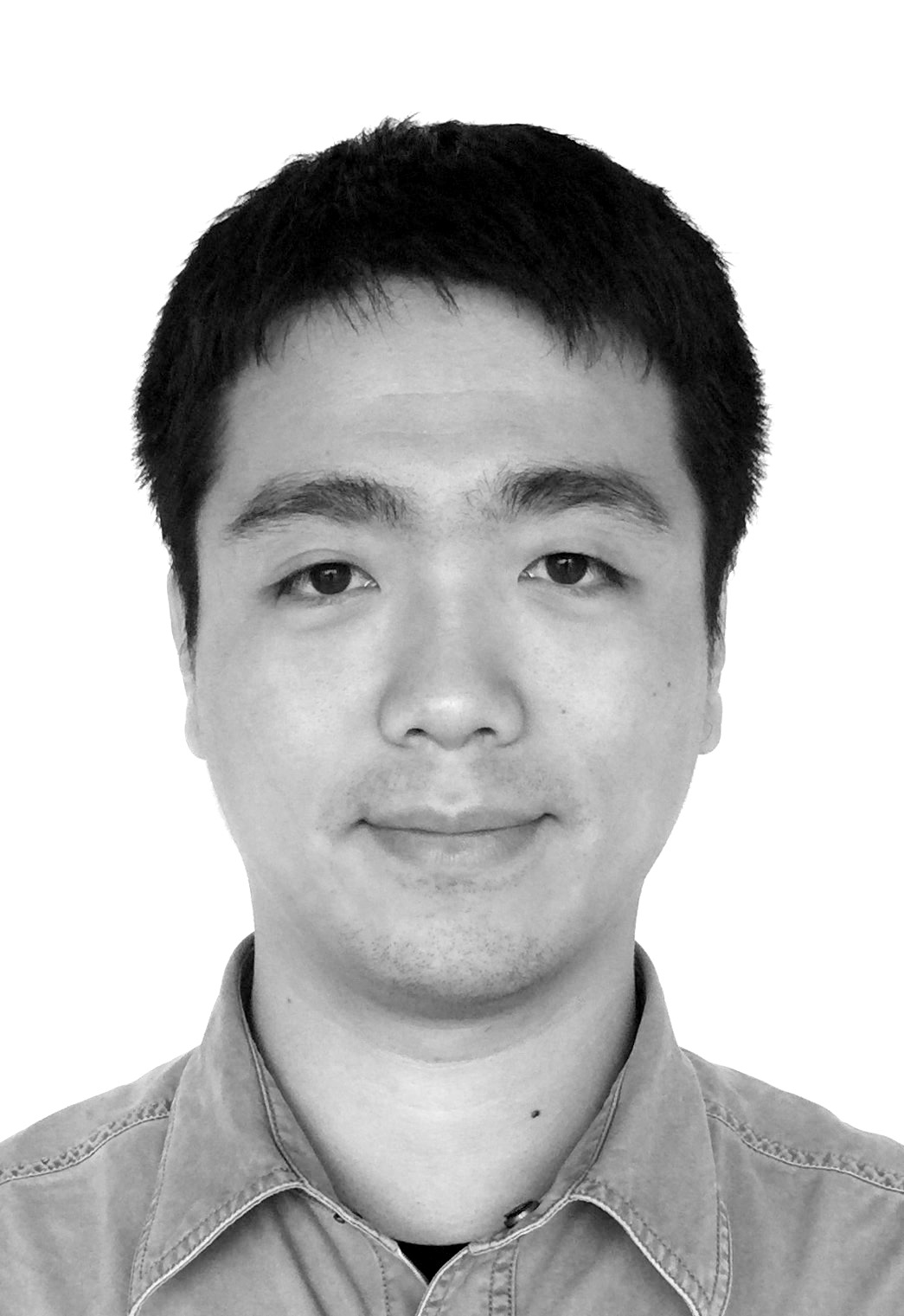}}]{Tian Huang}
    received his BSc and Ph.D. degree from the Shanghai Jiao Tong University, Shanghai, China in 2008 and 2016. He is currently a research scientist with Institute of High Performance Computing, Agency for Science Technology and Research, Singapore. His research interest lies in high-performance computing and machine learning algorithms.
\end{IEEEbiography}

\vfill
\newpage

\begin{IEEEbiography}[{\includegraphics[width=1in,height=1.25in,clip,keepaspectratio]{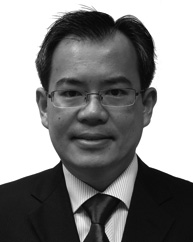}}]{Rick Siow Mong Goh}
    received the Ph.D. degree in electrical and computer engineering from the National University of Singapore, Singapore.

    He is the Director of the Computing \& Intelligence (CI) Department, Institute of High Performance Computing, Agency for Science, Technology and Research, Singapore, where he leads a team of over 80 scientists in performing world-leading scientific research, developing technology to commercialization, and engaging and collaborating with industry. His current research interests include artificial intelligence, high-performance computing, block chain, and federated learning.
\end{IEEEbiography}

\begin{IEEEbiography}[{\includegraphics[width=1in,height=1.25in,clip,keepaspectratio]{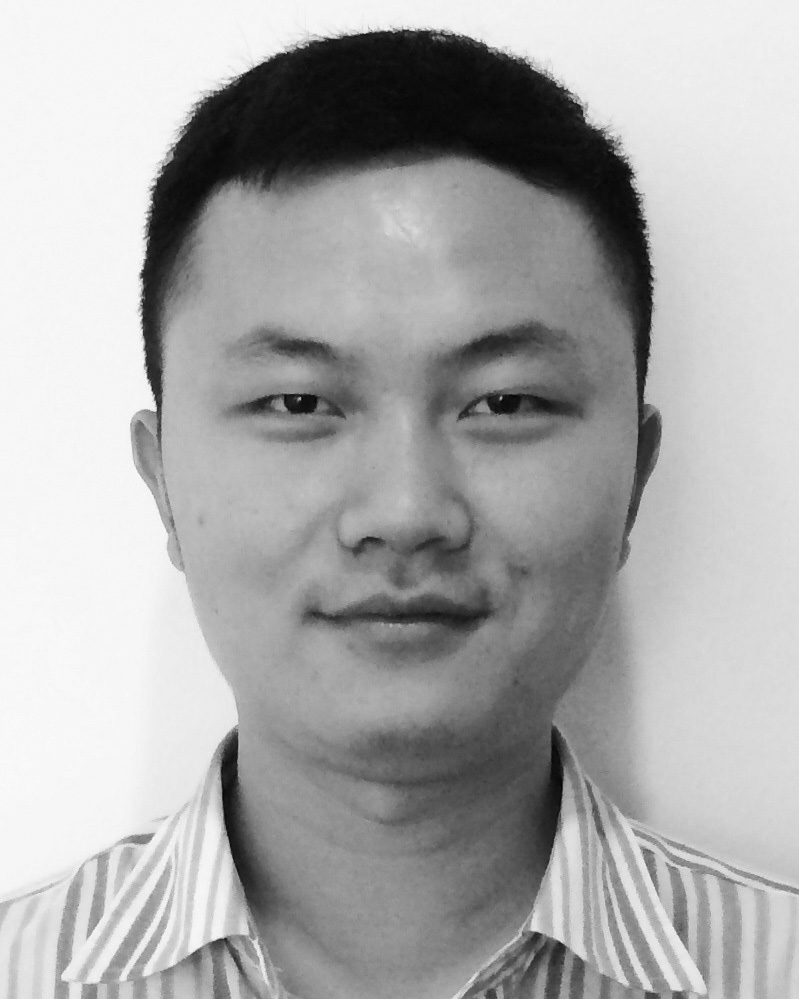}}]{Tao Luo}
    received his Bachelor degree in 2010 from Harbin Institute of Technology, China (HIT),  Master degree in 2013 from University of Electronic Science and Technology of China, China (UESTC), and Ph.D. in 2018 from School of Computer Science and Engineering of Nanyang Technological University, Singapore (NTU).
    
    He is currently a research scientist in Institute of High Performance Computing (IHPC), Agency for Science, Technology and Research, Singapore (A*STAR). His current research interests include Efficient AI, AI application, Neuromorphic Computing, High Performance Computing (HPC), and Reconfigurable Computing system.
\end{IEEEbiography}

\begin{IEEEbiography}[{\includegraphics[width=1in,height=1.25in,clip,keepaspectratio]{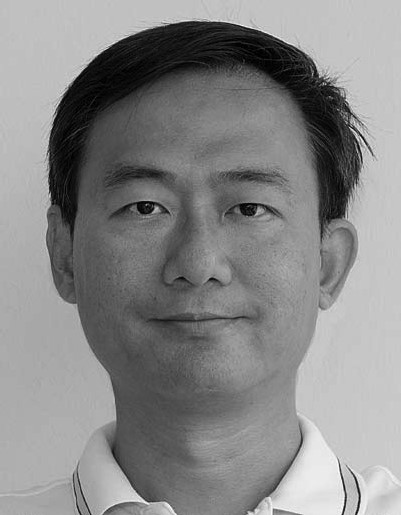}}]{Weng-Fai Wong}
received the BSc degree from the National University of Singapore, in 1988, and the DrEngSc degree from University of Tsukuba, Japan, in 1993. He is currently an associate professor at the Department of Computer Science, National University of Singapore. His research interests include computer architecture, compilers, and high-performance computing. He is a senior member of the IEEE.
\end{IEEEbiography}

\vfill

\end{document}